\documentclass[conference]{IEEEtran}
\IEEEoverridecommandlockouts
\usepackage[dvipsnames]{xcolor}
\usepackage{cite}
\usepackage{amsmath,amssymb,amsfonts}
\usepackage{graphicx}
\usepackage{multirow}
\usepackage{textcomp}
\usepackage{xcolor}
\usepackage{todonotes}
\usepackage{hyperref}
\usepackage{booktabs}
\usepackage{wrapfig}
\usepackage[numbers]{natbib}
\usepackage{subcaption}
\usepackage{algorithm}
\usepackage[noend]{algpseudocode}
\usepackage{cancel}   
\algrenewcommand\algorithmicindent{1.0em}%

\usepackage[frozencache,cachedir=.]{minted}
\usepackage{tcolorbox}
\definecolor{mintedbg}{RGB}{22,22,22}
\definecolor{mintedfg}{RGB}{248,248,242}
\newtcolorbox{mintedtcbox}{colback=mintedbg,fontupper=\color{mintedfg},hbox}
\RecustomVerbatimEnvironment{Verbatim}{BVerbatim}{}
\setminted{
    framesep=6pt,
    fontsize=\normalsize,
}
\newmintedfile[pythoncode]{python}{}
\newmintedfile[bashcode]{bash}{}
\usepackage{inconsolata}

\definecolor{mydarkblue}{rgb}{0,0.08,0.45}
\hypersetup{colorlinks,linkcolor=red,urlcolor=mydarkblue,linktoc=page}

\def\BibTeX{{\rm B\kern-.05em{\sc i\kern-.025em b}\kern-.08em
    T\kern-.1667em\lower.7ex\hbox{E}\kern-.125emX}}
\begin{document}

\title{A2C is a special case of PPO}


\author{\IEEEauthorblockN{Shengyi Huang}
\IEEEauthorblockA{
\textit{Drexel University}\\
USA \\
sh3397@drexel.edu}
\and
\IEEEauthorblockN{Anssi Kanervisto}
\IEEEauthorblockA{
\textit{University of Eastern Finland}\\
Finland \\
anssk@uef.fi}
\and
\IEEEauthorblockN{Antonin Raffin}
\IEEEauthorblockA{ 
\textit{German Aerospace Center}\\
Germany \\
antonin.raffin@dlr.de}
\and
\IEEEauthorblockN{Weixun Wang}
\IEEEauthorblockA{
\textit{Tianjin University}\\
China \\
wxwang@tju.edu.cn}
\and
\IEEEauthorblockN{Santiago Onta\~{n}\'{o}n$^{*}$\thanks{$^{*}$\text{Currently at Google}}}
\IEEEauthorblockA{
\textit{Drexel University}\\
USA \\
so367@drexel.edu}
\and
\IEEEauthorblockN{Rousslan Fernand Julien Dossa}
\IEEEauthorblockA{
\textit{Graduate School of System Informatics}\\
Kobe University \\
doss@ai.cs.kobe-u.ac.jp}
}

\IEEEoverridecommandlockouts

\maketitle

\begin{abstract}
Advantage Actor-critic (A2C) and Proximal Policy Optimization (PPO) are popular deep reinforcement learning algorithms used for game AI in recent years. A common understanding is that A2C and PPO are separate algorithms because PPO's clipped objective appears significantly different than A2C's objective. In this paper, however, we show A2C is a special case of PPO. 
We present theoretical justifications and pseudocode analysis to demonstrate why.
To validate our claim, we conduct an empirical experiment using \texttt{Stable-baselines3}, showing A2C and PPO produce the \textit{exact} same models when other settings are controlled.
\end{abstract}


\section{Introduction}

A2C~\citep{schulman2017proximal,mnih2016asynchronous} and PPO~\citep{schulman2017proximal} are popular deep reinforcement learning (DRL) algorithms used to create game AI in recent years. Researchers have applied either one to a diverse sets of games, including arcade games~\citep{mnih2013playing,schulman2017proximal}, soccer~\citep{Kurach2020GoogleRF}, board games~\citep{8848063,Kristensen2020StrategiesFU}, and complex multi-player games such as Dota 2~\citep{berner2019dota}, so many reputable DRL libraries implement A2C and PPO, which makes it easier for Game AI practitioner to train autonomous agents in games~\citep{JMLR:v22:20-1364,liang2018rllib,d2020mushroomrl,fujita2021chainerrl}.


A common understanding is that A2C and PPO are separate algorithms because PPO's clipped objective and training paradigm appears different compared to A2C's objective. 
As a result, almost all DRL libraries have architecturally implemented A2C and PPO as distinct algorithms. 

In this paper, however, we show A2C is a special case of PPO. We provide theoretical justifications as well as an analysis on the pseudocode to demonstrate the conditions under which A2C and PPO are equivalent. To validate our claim, we conduct an empirical experiment using \texttt{Stable-baselines3}~\citep{JMLR:v22:20-1364}, empirically showing A2C and PPO produce the \textit{exact} same models when other settings are controlled.\footnote{See code at \url{https://github.com/vwxyzjn/a2c_is_a_special_case_of_ppo}}


Our results demonstrate that it is not necessary to have a separate implementation of A2C in DRL libraries: they just need to include PPO and support A2C via configurations, reducing the maintenance burden for A2C in DRL libraries. 
More importantly, we contribute a deeper understanding of PPO to the Game AI community, empowering us to view past work in different perspective: we now know past works that compare PPO and A2C are essentially comparing two sets of hyperparameters of PPO.
Finally, our work points out the shared parts between PPO and A2C, which makes it easier to understand and attribute PPO's improvement, a theme highlighted by recent work~\citep{engstrom2019implementation,andrychowicz2021what}.



\section{Theoretical Analysis}
\label{sec:theory}

Using notation from \cite{schulman2017proximal}, A2C maximizes the following policy objective
\[L^{A2C}(\theta)=\hat{\mathbb{E}}_{t}\left[\log \pi_{\theta}\left(a_{t} \mid s_{t}\right) \hat{A}_{t}\right] ,\]
where $\pi_\theta$ is a stochastic policy parameterized by $\theta$, $\hat{A}_{t}$ is an estimator of the advantage function at timestep $t$,  $\hat{\mathbb{E}}_{t}[\ldots]$ is the expectation indicating the empirical average over a finite batch of samples, in an algorithm that alternates between sampling and optimization. When taking the gradient of the objective w.r.t. $\theta$, we get 
\[\nabla_\theta L^{A2C}(\theta)=\hat{\mathbb{E}}_{t}\left[\nabla_\theta \log \pi_{\theta}\left(a_{t} \mid s_{t}\right) \hat{A}_{t}\right] \]
In comparison, PPO maximizes the following policy objective \cite{schulman2017proximal}:
\begin{align*}
    L^{PPO}(\theta)=\hat{\mathbb{E}}_{t}\left[\min \left(r_{t}(\theta) \hat{A}_{t}, \operatorname{clip}\left(r_{t}(\theta), 1-\epsilon, 1+\epsilon\right) \hat{A}_{t}\right)\right]\\
    \text{where }r_t(\theta)= \frac{\pi_{\theta}\left(a_{t} \mid s_{t}\right)}{\pi_{\theta_{\text {old }}}\left(a_{t} \mid s_{t}\right)} \hat{A}_{t}
\end{align*}
At a first glance, $L^{A2C}$ appears drastically different from $L^{PPO}$ because 1) the $\log$ term disappeared in $L^{PPO}$, and 2) $L^{PPO}$ has clipping but $L^{A2C}$ does not.

\begin{tiny}
\begin{algorithm}[t]
\caption{Proximal Policy Optimization}\label{alg:ppo}
\begin{algorithmic}[1]
\State \textbf{Initialize} vectorized environment $E$ containing $N$ parallel environments
\State \textbf{Initialize} policy parameters $\theta_{\pi}$
\State \textbf{Initialize} value function parameters $\theta_{v}$
\State {\color{NavyBlue} \textbf{Initialize} Adam optimizer $O$ for $\theta_{\pi}$ and $\theta_{v}$ }
\State \textbf{Initialize} next observation $o_{next} = E.reset()$
\State \textbf{Initialize} next done flag $d_{next} = [0, 0, ..., 0]$ \# length $N$
\State
\For {$i$ = 0,1,2,..., $I$}
\State {(optional) \textbf{Anneal} learning rate $\alpha$ linearly to 0 with $i$}
\State \textbf{Set} $\mathcal{D} = (o, a, \log \pi (a|o), r, d, v)$ as tuple of 2D arrays
\State 
\State \# Rollout Phase:
\For {{\color{NavyBlue} $t$ = 0,1,2,..., $M$}}
\State \textbf{Cache} $o_{t} = o_{next}$ and $d_{t} = d_{next}$
\State \textbf{Get} $a_{t} \sim  \pi(\cdot| o_t)$ and $v_t = v(o_t)$
\State \textbf{Step} simulator: $o_{next}, r_t, d_{next} = E.step(a_t)$
\State \textbf{Let} $\mathcal{D}.o[t] = o_t, \mathcal{D}.d[t] = d_t, \mathcal{D}.v[t] = v_t, \mathcal{D}.a[t] = a_t$, $\mathcal{D}.\log \pi(a|o)[t] = \log \pi(a_t | o_t), \mathcal{D}.r[t] = r_t$
\EndFor
\State 
\State \# Learning Phase:
\State \textbf{Estimate / Bootstrap} next value $v_{next} = v(o_{next})$ 
\State {\color{NavyBlue}\textbf{Let} advantage $A = GAE(\mathcal{D}.r, \mathcal{D}.v, \mathcal{D}.d, v_{next}, d_{next}, \lambda)$}
\State \textbf{Let} $TD(\lambda)$ return $R = A + \mathcal{D}.v$
\State {\color{NavyBlue}  \textbf{Prepare} the batch $\mathcal{B} = {\mathcal{D}, A, R}$ and flatten $\mathcal{B}$}
\For {{\color{OliveGreen} $epoch$ = 0,1,2,..., $K$}}
\For {{\color{OliveGreen} mini-batch $\mathcal{M}$ of size $m$ in $\mathcal{B}$}}
\State {\color{OliveGreen} \textbf{Normalize} advantage $\mathcal{M}.A = \frac{\mathcal{M}.A - mean(\mathcal{M}.A)}{std(\mathcal{M}.A) + 10^{-8}}$}
\State {\color{NavyBlue} \textbf{Let} ratio $r = e^{\log \pi(\mathcal{M}.a | \mathcal{M}.o) - \mathcal{M}.\log \pi(a|o)}$}
\State {\color{NavyBlue} \textbf{Let} $ L^\pi=\min \left(r \mathcal{M}.A, \operatorname{clip}\left(r, 1-\epsilon, 1+\epsilon\right) \mathcal{M}.A\right)$}
\State {\color{NavyBlue} \textbf{Let} $L^V$ = clipped\_MSE$(\mathcal{M}.R, v(\mathcal{M}.o))$}
\State \textbf{Let} $L^S =  S[\pi(\mathcal{M}.o)]$
\State \textbf{Back-propagate} loss $L = -L^\pi + c_1 L^V - c_2 L^S $
\State \textbf{Clip} maximum gradient norm of $\theta_\pi$ and $\theta_v$ to $0.5$
\State \textbf{Step} the optimizer $O$ to initiate gradient descent
\EndFor
\EndFor
\EndFor
\end{algorithmic}
\end{algorithm}
\end{tiny}

\begin{tiny}
\begin{algorithm}[t]
\caption{Advantage Actor Critic}\label{alg:a2c}
\begin{algorithmic}[1]
\State \textbf{Initialize} vectorized environment $E$ containing $N$ parallel environments
\State \textbf{Initialize} policy parameters $\theta_{\pi}$
\State \textbf{Initialize} value function parameters $\theta_{v}$
\State {\color{NavyBlue} \textbf{Initialize} RMSprop optimizer $O$ for $\theta_{\pi}$ and $\theta_{v}$ }
\State \textbf{Initialize} next observation $o_{next} = E.reset()$
\State \textbf{Initialize} next done flag $d_{next} = [0, 0, ..., 0]$ \# length $N$
\State
\For {$iteration$ = 0,1,2,..., $I$}
\State {(optional) \textbf{Anneal} learning rate $\alpha$ linearly to 0 with $i$}
\State \textbf{Set} $\mathcal{D} = (o, a, \log \pi (a|o), r, d, v)$ as tuple of 2D arrays
\State 
\State \# Rollout Phase:
\For {{\color{NavyBlue} $t$ = 0,1,2,..., $M=5$}}
\State \textbf{Cache} $o_{t} = o_{next}$ and $d_{t} = d_{next}$
\State \textbf{Get} $a_{t} \sim  \pi(\cdot| o_t)$ and $v_t = v(o_t)$
\State \textbf{Step} simulator: $o_{next}, r_t, d_{next} = E.step(a_t)$
\State \textbf{Let} $\mathcal{D}.o[t] = o_t, \mathcal{D}.d[t] = d_t, \mathcal{D}.v[t] = v_t, \mathcal{D}.a[t] = a_t$, $\mathcal{D}.\log \pi(a|o)[t] = \log \pi(a_t | o_t), \mathcal{D}.r[t] = r_t$
\EndFor
\State 
\State \# Learning Phase:
\State \textbf{Estimate / Bootstrap} next value $v_{next} = v(o_{next})$ 
\State {\color{NavyBlue}\textbf{Let} advantage $A = GAE(\mathcal{D}.r, \mathcal{D}.v, \mathcal{D}.d, v_{next}, d_{next}, 1)$}
\State \textbf{Let} $TD(\lambda)$ return $R = A + \mathcal{D}.v$s
\State {\color{NavyBlue} \textbf{Prepare} the batch $\mathcal{M} = {\mathcal{D}, A, R}$ and flatten $\mathcal{M}$}
\State 
\State 
\State 
\State 
\State {\color{NavyBlue} \textbf{Let} $L^\pi= - \log \pi(o|a) M.A$}
\State {\color{NavyBlue} \textbf{Let} $L^V = (\mathcal{M}.R - v(\mathcal{M}.o))^2$}
\State \textbf{Let} $L^S =  S[\pi(\mathcal{M}.o)]$
\State \textbf{Back-propagate} loss $L = -L^\pi + c_1 L^V - c_2 L^S $
\State \textbf{Clip} maximum gradient norm of $\theta_\pi$ and $\theta_v$ to $0.5$
\State \textbf{Step} the optimizer $O$ to initiate gradient descent
\EndFor
\end{algorithmic}
\end{algorithm}
\end{tiny}

Nevertheless, note that $\pi_{\theta}$ and $\pi_{\theta_{\text {old }}}$ are the same during PPO's first update epoch, which means $r_t(\theta) = 1$. Hence, the clipping operation would not be triggered. Since no clipping happens, both terms in minimum operation are same, so the minimum operation does nothing. Thus if PPO sets the number of update epochs $K$ to 1, $L^{PPO}$ collapses into 
\begin{align*}
    L^{PPO, K=1}(\theta) & =\hat{\mathbb{E}}_{t}\left[\min \left( r_t(\theta) \hat{A}_{t}, r_t(\theta) \hat{A}_{t}\right)\right] \\
    &= \hat{\mathbb{E}}_{t}\left[r_t(\theta) \hat{A}_{t}\right] = \hat{\mathbb{E}}_{t}\left[\frac{\pi_{\theta}\left(a_{t} \mid s_{t}\right)}{\pi_{\theta_{\text {old }}}\left(a_{t} \mid s_{t}\right)} \hat{A}_{t}\right] 
\end{align*}
Now, we can take the gradient of this objective w.r.t $\theta$ and apply the ``log probability tricks" used in REINFORCE~\citep{sutton2018reinforcement,58756}:
\begin{align*}
    \begin{aligned} \nabla_{\theta} L^{PPO, K=1}(\theta) &=\nabla_{\theta} \hat{\mathbb{E}}_{t}\left[\frac{\pi_{\theta}\left(a_{t} \mid s_{t}\right)}{\pi_{\theta_{\text {old }}}\left(a_{t} \mid s_{t}\right)} \hat{A}_{t}\right] \\
        &=\hat{\mathbb{E}}_{t}\left[\frac{\nabla_{\theta} \pi_{\theta}\left(a_{t} \mid s_{t}\right)}{\pi_{\theta_{\text {old }}}\left(a_{t} \mid s_{t}\right)} \hat{A}_{t}\right] \\
        &=\hat{\mathbb{E}}_{t}\left[\frac{\pi_{\theta}\left(a_{t} \mid s_{t}\right)}{\pi_{\theta}\left(a_{t} \mid s_{t}\right)} \frac{\nabla_{\theta} \pi_{\theta}\left(a_{t} \mid s_{t}\right)}{\pi_{\theta_{\text {old }}}\left(a_{t} \mid s_{t}\right)} \hat{A}_{t}\right] \\
        &=\hat{\mathbb{E}}_{t}\left[\frac{\pi_{\theta}\left(a_{t} \mid s_{t}\right)}{\pi_{\theta_{\theta l d}}\left(a_{t} \mid s_{t}\right)} \frac{\nabla_{\theta}\pi_{\theta}\left(a_{t} \mid s_{t}\right)}{\pi_{\theta}\left(a_{t} \mid s_{t}\right)} \hat{A}_{t}\right] \\
        &=\hat{\mathbb{E}}_{t}\left[ \cancel{ \frac{\pi_{\theta}\left(a_{t} \mid s_{t}\right)}{\pi_{\theta_{\text {old }}}\left(a_{t} \mid s_{t}\right)} } \nabla_{\theta} \log \pi_{\theta}\left(a_{t} \mid s_{t}\right) \hat{A}_{t}\right]\\
        &=\hat{\mathbb{E}}_{t}\left[ \nabla_{\theta} \log \pi_{\theta}\left(a_{t} \mid s_{t}\right) \hat{A}_{t}\right]= \nabla_\theta L^{A2C}(\theta)
        \end{aligned}
\end{align*}
Note that when $K=1$, the ratio $\frac{\pi_{\theta}\left(a_{t} \mid s_{t}\right)}{\pi_{\theta_{\text {old }}}\left(a_{t} \mid s_{t}\right)}=1$ because $\pi_{\theta}$ has not been updated and is the same as $\pi_{\theta_{\text {old }}}$.
Therefore when $K=1$, PPO and A2C share the same gradient given the same data.

\section{Implementation Analysis}

PPO is an algorithm with many important implementation details~\citep{shengyi2022the37implementation,engstrom2019implementation,andrychowicz2021what}, so it can be challenging to see how our theoretical results relate to actual implementation.  To help make the connection between theory and implementations, we have prepared an complete pseudocode for PPO and A2C in Algorithm~\ref{alg:ppo} and \ref{alg:a2c}, respectively.


\begin{figure*}[t]
\centering
{ 
\scalebox{0.78}{
  \begin{mintedtcbox}
  \pythoncode{code/sb3_a2c.py}
  \end{mintedtcbox}
  }
 }
{
\hspace{-6mm}
\scalebox{0.75}{
  \begin{mintedtcbox}
  \pythoncode{code/sb3_ppo.py}
  \end{mintedtcbox}
  }}

\vspace{2mm}
\hfill
{\includegraphics[width=0.99\textwidth]{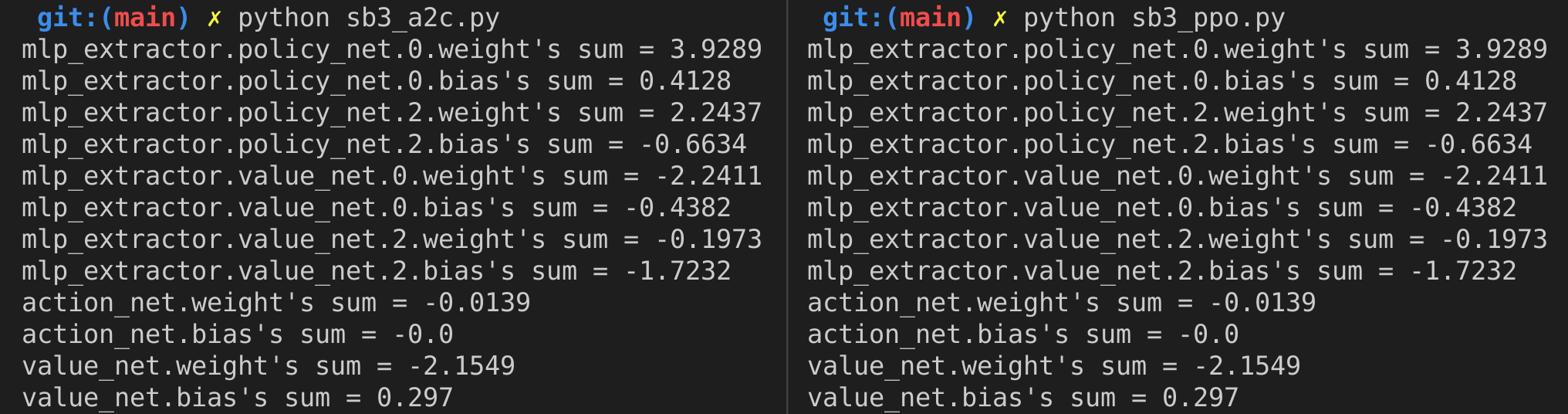}}\hfill
  \caption{The source code of our experiments and execution results. As shown, PPO produces the exact same trained model as A2C after aligning the settings.
  }
  \label{fig:sb3-code}
\end{figure*}

To highlight their differences, we labeled  {\color{OliveGreen} the code that \emph{only} PPO has with  green lines} and {\color{NavyBlue} the code that PPO and A2C differs with  blue lines}. As shown, the differences are that
\begin{enumerate}
    \item {\color{NavyBlue}PPO uses a different optimizer (line 4 in Algorithm~\ref{alg:ppo}).}
    \item {\color{NavyBlue}PPO modifies the number of steps $M$ in the rollout phase (line 13 in Algorithm~\ref{alg:ppo}).} PPO uses $M=128$ for Atari games and $M=2048$ for MuJoCo robotics tasks~\citep{schulman2017proximal}; in comparison, A2C consistently uses $M=5$, which is the corresponding hyperparameter $t_{max}=5$ in Asynchronous Advantage Actor Critic (A3C)~\citep{mnih2016asynchronous}.
    \item {\color{NavyBlue} PPO estimates the advantage with Generalized Advantage Estimation (GAE)~\citep{Schulman2016HighDimensionalCC} (line 21 in Algorithm~\ref{alg:ppo})}, whereas A2C estimates the advantage by subtracting the values of states from returns~\citep[page 3]{mnih2016asynchronous}, which is a special case of GAE when $\lambda = 1$~\citep[Equation 18]{Schulman2016HighDimensionalCC}.
    \item  {\color{OliveGreen}PPO does gradient updates on the rollout data for $K$ epochs~\citep[page 5]{schulman2017proximal} each with mini-batches~\citep[see ``6. Mini-batch Updates'']{shengyi2022the37implementation} (line 23-25 in Algorithm~\ref{alg:ppo})}, whereas A2C just does a single gradient update on the entire batch of rollout data (line 23 in Algorithm~\ref{alg:a2c}).
    \item {\color{OliveGreen} PPO normalizes the advantage (line 26 in Algorithm~\ref{alg:ppo})}.
    \item {\color{NavyBlue} PPO uses the the clipped surrogate objective (line 27-28 in Algorithm~\ref{alg:ppo})}, which we showed in section~\ref{sec:theory} is equivalent to A2C's objective when $K = 1$ and $|B| = |M|$ (i.e., not splitting data to mini-batches). 
    \item {\color{NavyBlue} PPO uses a clipped mean squared error loss (line 29 in Algorithm~\ref{alg:ppo})}~\citep[see ``9. Value Function Loss Clipping'']{shengyi2022the37implementation}, where as A2C just uses the regular mean squared error loss.
\end{enumerate}

To show A2C is a special case of PPO, we would need to remove the green code and align settings with the blue code, as shown in the following section.

\section{Empirical Experiments}

To validate our claim, we conduct an experiment with \texttt{CartPole-v1} using the A2C and PPO models implemented in \texttt{Stable-baselines3}~\citep{JMLR:v22:20-1364}. Specifically, we made the following configurations to PPO:
\begin{enumerate}
    \item {\color{NavyBlue} Match A2C's RMSprop optimizer and configurations (i.e., $\alpha=0.99, \epsilon=0.00001$, and zero for weight decay) set the learning rate $\alpha=0.0007$ (also means turning off learning rate annealing).}
    \item {Match the learning rate to be $0.0007$, which also disables the learning rate annealing.
    }
    \item {\color{NavyBlue} Match the number of steps $M$ parameter to be $5$.}
    \item {\color{NavyBlue} Disable GAE by setting its $\lambda= 1$}
    \item {\color{OliveGreen} Set the number of update epochs $K$ to 1, so the clipped objective has nothing to clip; also, perform gradient update on the whole batch of training data (i.e., do not use mini-batches\footnote{in \texttt{Stable-Baselines3} we set the \texttt{batch\_size = num\_envs * num\_steps}. In \texttt{openai/baselines} we set \texttt{nminibatches}, the number of mini-batches, to 1.}).}
    \item {\color{OliveGreen} Turn off advantage normalization.}
    \item {\color{NavyBlue}  Disable value function clipping.
    }
\end{enumerate}

Figure~\ref{fig:sb3-code} shows the source code and results. After 3000 steps of training, we see A2C and PPO produce the exact same trained model after properly setting random seeds in all dependencies. Hence, A2C is hence a special case of PPO when aligning configurations as shown above.


\section{Conclusion}
In this paper, we demonstrated A2C is a special case of PPO. We first provide theoretical justification that explains how PPO's objective collapses into A2C's objective when PPO's number of update epochs $K$ is 1. Then we conducted empirical experiments via \texttt{Stable-baselines3} to show A2C and PPO produce the exact same model when other settings and all sources of stochasticity are controlled. With this insight, Game AI practitioners can implement a single core code for PPO and A2C, reducing maintenance burden. Furthermore, given the wide adoption of A2C and PPO in training Game AI, our work contributes a deeper understanding of how A2C and PPO are related.

\bibliographystyle{IEEEtran}
\bibliography{references}

\clearpage
\onecolumn

\end{document}